\documentclass[10pt,twocolumn,letterpaper]{article}

\usepackage{cvpr}              

\usepackage[dvipsnames]{xcolor}
\usepackage{times}
\usepackage{epsfig}
\usepackage{graphicx}
\usepackage{amsmath}
\usepackage{amssymb}
\usepackage{algorithm}
\usepackage{algorithmic}
\usepackage{comment}
\usepackage{xcolor}
\usepackage{multirow}
\usepackage{multicol}
\usepackage{color, colortbl}
\usepackage[accsupp]{axessibility}  

\DeclareUnicodeCharacter{0301}{\'{e}}

\definecolor{cvprblue}{rgb}{0.21,0.49,0.74}
\usepackage[pagebackref,breaklinks,colorlinks,citecolor=cvprblue]{hyperref}

\newcommand*\samethankss[1][\value{footnote}]{\footnotemark[#1]}
\def\paperID{2592} 
\def\confName{CVPR}
\def\confYear{2025}

\title{Homogeneous Dynamics Space for Heterogeneous Humans}

\author{
Xinpeng Liu$^{1,2}$,~
Junxuan Liang$^1$,~
Chenshuo Zhang$^1$,~
Zixuan Cai$^3$,~
Cewu Lu$^{1,2}$\thanks{Corresponding authors.},~
Yong-Lu Li$^{1,2}$\samethankss~\\
\tt\small{$^1$Shanghai Jiao Tong University}, \tt\small{$^2$Shanghai Innovation Institute}, \tt\small{$^3$Soochow University}\\
\tt\small{xinpengliu0907@gmail.com}, \tt\small{zxcai@stu.suda.edu.cn}, \\
\tt\small{\{whitefork, zhangchenshuo, lucewu, yonglu\_li\}@sjtu.edu.cn}\\
}

\begin{document}
\maketitle
\begin{abstract}
Analyses of human motion kinematics have achieved tremendous advances.
However, the production mechanism, known as human dynamics, is still undercovered.
In this paper, we aim to push the understanding of data-driven human dynamics forward.
We identify a major obstacle to this as the \textbf{heterogeneity} of existing human motion understanding efforts.
Specifically, heterogeneity exists in not only the diverse kinematics representations and hierarchical dynamics representations but also the data from different domains, namely biomechanics and reinforcement learning.
With an in-depth analysis of the existing heterogeneity, we propose to emphasize the beneath homogeneity: all of them represent the \textbf{homogeneous} fact of human motion, though from different perspectives.
Given this, we propose \textbf{H}omogeneous \textbf{Dy}namics \textbf{S}pace (HDyS) as a fundamental space for human dynamics by aggregating heterogeneous data and training a homogeneous latent space with inspiration from the inverse-forward dynamics procedure.
HDyS achieves decent mapping between human kinematics and dynamics by leveraging the heterogeneous representations and datasets.
We demonstrate the feasibility of HDyS with extensive experiments and applications.
The project page is \href{https://foruck.github.io/HDyS}{https://foruck.github.io/HDyS}.
\end{abstract}

\section{Introduction}

Analyses on human motion have a wide range of applications, including animation~\cite{guo2022generating,tevet2022human}, healthcare~\cite{gordon2023learning,sedighi2023emg}, and robotics~\cite{skuric2022line,yaacoub2023probabilistic}. 
The computer vision community has made tremendous progress in understanding human kinematics with tasks like human reconstruction~\cite{li2021hybrik,djrn,gio}, action recognition~\cite{tincvpr,punnakkal2021babel,pastanet,li2024sadvaeimprovingzeroshotskeletonbased,hakev2}, and motion generation~\cite{tevet2022human,liu2024revisithumansceneinteractionspace,pangea}.
However, the production mechanism hidden beneath human motion, known as human dynamics, is still limitedly explored. 

In this paper, we aim to push the understanding of human dynamics forward with data-driven methodologies. 
Specifically, we pursue to build a bidirectional mapping between human kinematics and dynamics.
We identify the major obstacle to this as the heterogeneity of existing human motion understanding efforts, which could be two-fold.

\begin{figure}[!t]
    \centering
    \includegraphics[width=\linewidth]{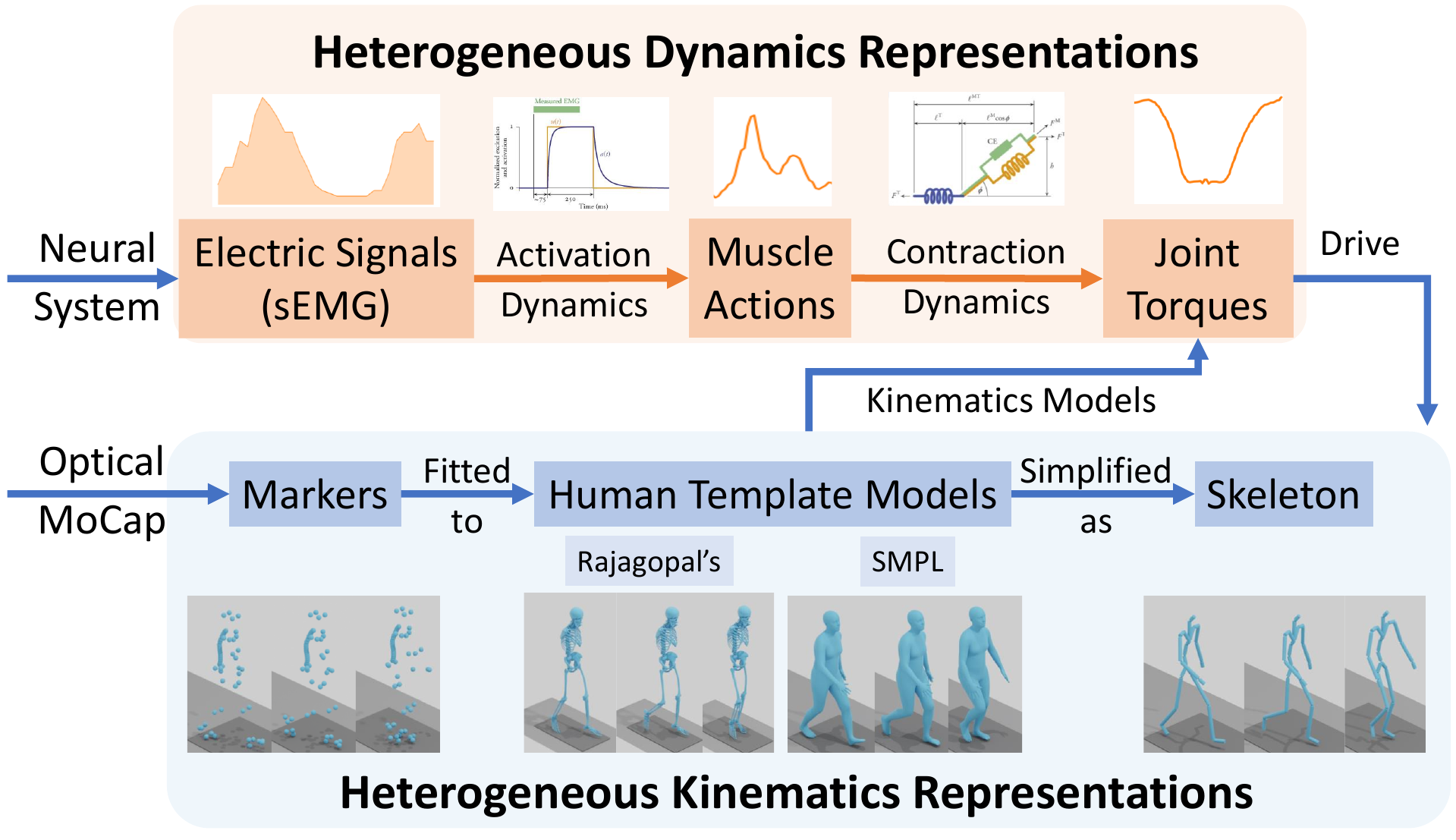}
    \vspace{-10px}
    \caption{The representation heterogeneity exists for both kinematics and dynamics.}
    \vspace{-10px}
    \label{fig:teaser-hetero}
\end{figure}

First, \textbf{representation heterogeneity} exists for both kinematics and dynamics, as shown in Figure~\ref{fig:teaser-hetero}.
A typical optical motion capture pipeline, which produces most existing available data for human kinematics~\cite{mahmood2019amass,skel}, involves tracking the \textit{markers} placed on the surface of humans. 
Then, different human template models are fitted to the tracked markers, resulting in \textit{parameters} of musculoskeletal models~\cite{rajagopal2016full} for biomechanics uses or SMPL-like models~\cite{loper2023smpl} for general CV/CG uses.
\textit{Skeletons} with pre-defined kinematic trees are sometimes preferred as a simple and intuitive representation.
Different datasets tend to include only representations preferred by their designed goal.
Besides the kinematics representation heterogeneity, the dynamics representation heterogeneity exists more severely.
Human dynamics function in a hierarchical manner.
The closest dynamics hierarchy to the kinematics is the \textit{joint torques}, which are further produced by muscle forces from \textit{muscle actions} according to the muscle contraction dynamics.
Furthermore, \textit{electric signals} from the neural system activate the muscle, which surface electromyography (sEMG) could detect.
Effective conversions between these different representations are not always available.
For example, for kinematics representations, the conversion between the musculoskeletal model and SMPL becomes available in very recent advances~\cite{skel}, which requires time-consuming optimization.
For dynamics, the bi-directional mapping between joint torques, muscle actions, and electric signals is even more difficult.

Second, \textbf{domain heterogeneity} also exists.
One crucial characteristic of human dynamics is the difficulty of non-intrusive measurement.
In the biomechanics community, optimization-based methods are adopted to solve the Euler-Newton equation for joint torques~\cite{werling2023addbiomechanics} and muscle actions~\cite{schneider2024mint}.
For electric signals, it is typically measured by sEMG sensors~\cite{chiquier2023muscles}.
In the meantime, from the learning community, there emerge efforts~\cite{liu2024imdy} utilizing Reinforcement Learning (RL) to imitate existing motion captures with physically simulated humanoids and record the simulated human dynamics data in a fully synthetic manner.
These different data sources introduce domain heterogeneity in three aspects.
First, biomechanics data are more deeply rooted in reality with well-defined computation models and real human motions.
Instead, RL data could diverge from real-world situations with unnatural motion imitations and inaccurate physics simulations, known as the sim2real domain gap.
Second, biomechanics data, due to the optimization-based manner, is sensitive to variations and usually limited to strictly controlled laboratory setups with relatively simple motions, resulting in limited motion domain coverage.
In contrast, RL data cover a wide range of motion with more diversity.
Third, biomechanics data and RL data tend to adopt heterogeneous kinematics representations with different kinematic trees, making it hard to transfer directly from one domain to another.

Despite the heterogeneity, we emphasize the homogeneous essence behind it: all of them represent the homogeneous fact of human motion, each from a different perspective.
Given this, we demonstrate the feasibility of unifying these heterogeneous human representations and exploiting the underlying homogeneous knowledge of human dynamics.
Low-level Cartesian kinematics representations, like markers and joints, are less heterogeneous than representations like joint angles. 
Also, though joint torques, muscle actions, and electric signals are not directly transferable to each other, they could share similar motor knowledge like coordination.
To this end, we propose \textbf{H}omogeneous \textbf{Dy}namics \textbf{S}pace for heterogeneous humans (HDyS).

To achieve this, we aggregate human dynamics data from both RL~\cite{liu2024imdy} and biomechanics~\cite{werling2024addbiomechanics,chiquier2023muscles,schneider2024mint} with different kinematics and dynamics representations, covering human dynamics hierarchies, including joint torques, muscle actions, and electric signals.
Then, HDyS is designed as an aggregation of multiple auto-encoders corresponding to the inverse-forward dynamics procedure.
We supervise HDyS with reconstruction and alignment losses.
Our proposed HDyS has two major merits.
First, by unifying heterogeneous human representation in the same latent space, it generalizes across different representations seamlessly while taking advantage of each.
Second, by aggregating large-scale heterogeneous motion data, it is empowered with general human motor knowledge from a wide span of motion, functioning as a reusable knowledge source for downstream dynamics-related applications.
To demonstrate the efficacy of HDyS, we first evaluate it on human inverse dynamics.
Then, we showcase how HDyS could facilitate downstream dynamics-related applications like ground reaction force prediction, biomechanics human simulation, and physics-simulated character control.

To summarize, our contribution includes: 
1) We analyzed the heterogeneity issue that hinders an in-depth understanding of human dynamics.
2) We highlighted the homogeneity beneath the heterogeneity and proposed a fundamental reusable space HDyS for human dynamics by unifying the heterogeneity.
3) We demonstrated the feasibility of digging homogeneity out from heterogeneity with extensive experiments and applications of HDyS.
\section{Related Works} 

\subsection{Human Dynamics} 
By human dynamics, we mean the production mechanism of human motion, which the biomechanics community has actively explored.
To produce a certain motion, neural commands are sent to activate the muscles.
After receiving the activation signals, muscles contract and produce muscle forces.
Multiple muscle forces form the joint torques according to certain musculoskeletal geometry, and the joint torques drive the accelerations that accumulate into movements. 
Thus, understanding human dynamics typically involves two heterogeneous hierarchies: joint torques and muscle activations.
However, both are hard to measure non-intrusively. 
In the literature, to obtain them, an optimization problem is typically introduced as
\begin{equation}
    \begin{aligned}
        \min \|a\|, s.t.& \ 0 \leq a \leq 1, \tau=A(q)F(a),
        \\& M(q)\ddot{q} + C(q,\dot{q}) + G(q) = J\lambda + \tau.
    \end{aligned}
    \label{eq:1}
\end{equation}
with the generalized human inertia matrix $M(q)$ w.r.t. generalized coordinate $q$, Coriolis and centrifugal forces $C(q,\dot{q})$, gravity $G(q)$, Jacobian matrix $J$ mapping external forces $\lambda$ to the generalized coordinates, and muscle activations $a$.
The joint torques $\tau$ could be obtained by $\tau=A(q)F(a)$, where $A(q)$ maps muscle forces into joint torques and $F(a)$ maps $a$ into muscle forces usually with the hill-type function~\cite{hill1938heat,zajac1989muscle}.
Mature software~\cite{delp2007opensim,damsgaard2006analysis,werling2021fast} were developed for this purpose.
However, the optimization quality is tightly bonded to the precision of external force measurement, which could be expensive.
Therefore, the applications are mostly limited to simple motions like gaits in laboratory settings. 
Some efforts exploited wearable devices~\cite{Latella2016,latella2019simultaneous} for more general applications.
In addition, the optimization is deeply coupled with the adopted human models~\cite{rajagopal2016full,loper2023smpl}, which vary with the application preferences. 
Fitting raw motion capture data to a specific human model could be unstable and time-consuming.
The conversion between different human models is also non-trivial even with recent advances~\cite{skel}.
This way, transferring dynamics from one human model to another is limitedly explored.
Besides torques and muscle actions that are typically obtained via optimization, Surface Electromyography (sEMG) is adopted as an indirect representation of human dynamics which could be directly measured. 
Efforts were made to build the accurate mapping from sEMG to muscle actions~\cite{kang2020event,wimalasena2022estimating}, joint torques~\cite{schulte2022multi,zhang2023estimation}, and human poses~\cite{liu2021neuropose,sedighi2023emg}.
Though progress was made, sEMG patterns might suffer from noises and vary drastically among subjects~\cite{trepman1998electromyographic}, hindering the generalization. 

\subsection{Learning-based Human Dynamics}

\noindent{\bf Joint Torques.}
Early efforts were made on ML-based joint torque analysis for certain human body parts~\cite{johnson2014efficient,xiong2019intelligent,manukian2023artificial}.
Lv~\etal~\cite{lv2016data} developed a Gaussian mixture framework for whole-body joint torque estimation.
Other architectures like k-nearest neighbor-based regression~\cite{zell2015physics,zell2017joint}, random forests~\cite{zell2017learning}, and neural networks~\cite{zell2020weakly} were also adopted for the estimation of joint torques.
However, most of these efforts suffered from limited data scale, which hampered learning methods from exploiting their full potential.
The recent emergence of AddBiomechanics~\cite{werling2024addbiomechanics}, which aggregated multiple biomechanics datasets, considerably boosted the data scale. 
However, most of the collected sequences contained only gaits with limited diversity. 
Another line of work adopted reinforcement learning to simulate motion in physics simulators, and the joint torques could be obtained in simulation.
Though generalizability was limited at first~\cite{bergamin2019drecon,peng2021amp,won2021control,won2022physics,peng2022ase}, emerging efforts~\cite{yuan2019ego,luo2021dynamics,luo2023perpetual} managed to replicate a wide span of motion in simulators. 
The paradigm is further incorporated into recent MoCap systems for simultaneous estimation of motion and the joint torques~\cite{yi2022physical,gartner2022differentiable,gartner2022trajectory,huang2022neural,wang2023learning,zhang2024incorporating}.
However, due to the involvement of simulators, the sim2real gap exists.
The learned dynamics could be restricted to certain simulators or human models, which might diverge from real humans.

\noindent{\bf Muscle Actions.}
sEMG is usually adopted as a proxy measurement of muscle actions and has been extensively studied in biomechanics~\cite{feldotto2022evaluating,hernandez2023kinematic,moreira2021lower,furmanek2022kinematic,malevsevic2021database,chiquier2023muscles,peng2023musclemap}. 
There have been efforts delving into predicting sEMG signals given either joint torques~\cite{sekiya2019linear,song2015inverse,li2014inverse}, goniometers~\cite{tamilselvam2021musculoskeletal}, motion captures~\cite{johnson2009evaluation,yamane2009muscle,nakamura2005somatosensory}, videos~\cite{chiquier2023muscles,peng2023musclemap}, or point clouds~\cite{niu2022estimating}.
However, each effort could be small in data scale, motion variations, and muscle coverage. 
Unifying these efforts could be hindered by the heterogeneous settings and data formats adopted, thus under-covered.
Recently, musculoskeletal human simulation has gained attention.
Multiple efficient simulators were developed~\cite{caggiano2022myosuite,Geijtenbeek2021Hyfydy,zuo2024self}, opening the potential for more accessible muscle action analysis.
Specifically, MinT~\cite{schneider2024mint} was proposed by simulating motion sequences from AMASS~\cite{mahmood2019amass} in OpenSim~\cite{delp2007opensim} and attaching simulated muscle actions to the raw motion sequence, enabling muscle action learning in scale.

\section{Method}

We introduce the proposed Homogeneous Dynamics Space (HDyS). 
We first cover the involved kinematics and dynamics representations in Section~\ref{sec:kin-rep},~\ref{sec:dyn-rep}.
Then, the model architecture and designed losses are introduced in Section~\ref{sec:hdys}.

\subsection{Kinematics Representations}
\label{sec:kin-rep}
\begin{figure}[!t]
    \centering
    \includegraphics[width=\linewidth]{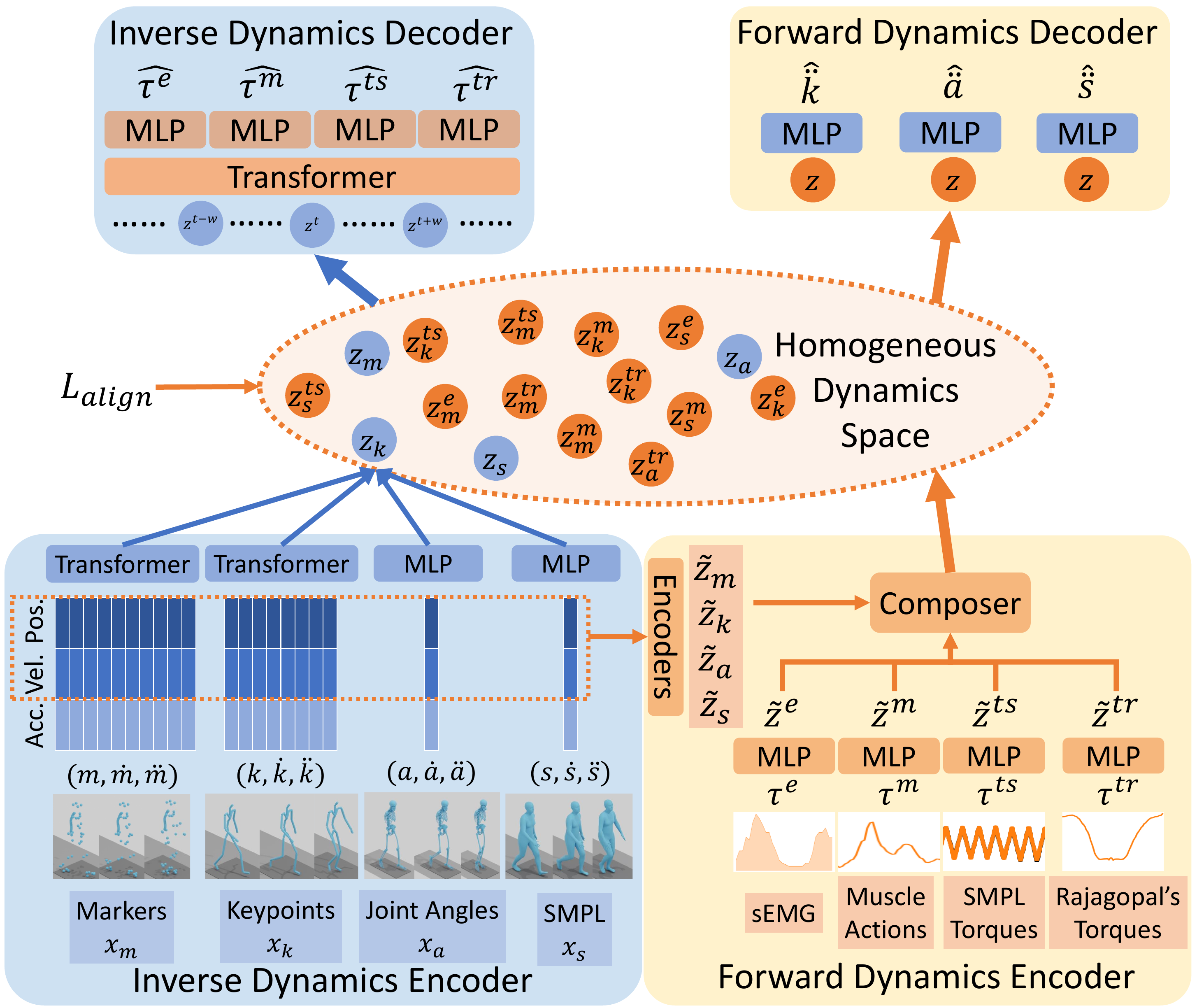} 
    \vspace{-10px}
    \caption{The overall architecture of HDyS.}
    \vspace{-15px}
    \label{fig:pipe}
\end{figure}

As shown in Fig.~\ref{fig:pipe}, we adopt four types of kinematics representations: Cartesian representations of markers and skeleton key-points, joint angles of Rajagopal's model~\cite{rajagopal2016full} from the biomechanics community, and SMPL~\cite{loper2023smpl} which is widely adopted for CV and CG applications.

\noindent{\bf Markers}, placed on the human body surfaces, are typically the raw data for optical motion capture.
In practice, most other representations are calculated by fitting certain human prior models to the marker observations, making it easy to obtain for heterogeneous datasets and representations.
Thus, markers are expected to be a generalizable representation across heterogeneous datasets, though they could also be rather low-level and thus hard to learn.
We define the marker representation at timestamp $t$ as $x_m^t=(m^t, \dot{m}^t, \dot{m}^) \in \mathcal{R}^{N_m \times 9}$, which is composed of marker Cartesian coordinates $m^t$, finite-differentiated velocities $\dot{m}^t$ and finite-differentiated accelerations $\dot{m}^t$ with $N_m$ markers.

\noindent{\bf Skeletal Keypoints}, compared to markers, are less generalizable due to their reliance on pre-defined kinematic trees. 
However, with its Cartesian coordinates, common sense of human topology, and easy access, the gap between different kinematic trees can be mitigated.
We define the joint representation at timestamp $t$ as $x_k^t=(k^t, \dot{k}^t, \dot{k}^) \in \mathcal{R}^{N_k \times 9}$, which is composed of joint coordinates $k^t$, finite-differentiated velocities $\dot{k}^t$ and finite-differentiated accelerations $\dot{k}^t$ with $N_k$ skeletal joints.

\noindent{\bf Joint Angles} are preferred for clinical analyses which more faithfully preserves the biomechanics information of human kinematics.
We adopt the Rajagopal's model~\cite{rajagopal2016full} used in AddBiomechanics~\cite{werling2024addbiomechanics}, and define the joint angle representation at timestamp $t$ as $x_a^t=(a^t, \dot{a}^t, \dot{a}^) \in \mathcal{R}^{3N_j}$, containing joint angles $a^t$, finite-differentiated velocities $\dot{a}^t$ and accelerations $\dot{a}^t$ with $N_j$ joints in Rajagopal's model (only 23 lower-body joints are used).

\noindent{\bf SMPL~\cite{loper2023smpl}} is a human parameter model widely adopted in the CV/CG community. 
We define the SMPL representation at timestamp $t$ as $x_s^t=(s^t, \dot{s}^t, \dot{s}^) \in \mathcal{R}^{3N_j}$, which is composed of SMPL parameters $s^t$, finite-differentiated velocities $\dot{s}^t$ and finite-differentiated accelerations $\dot{s}^t$ with $N_j$ joints of SMPL (75 used, 3 translational joints at root and 72 revolving joints).

\subsection{Dynamics Representations}
\label{sec:dyn-rep}
We adopt three different types of dynamic representations: joint torques, muscle actions, and sEMGs.

\noindent{\bf Joint Torques} are the net torques exerted at each joint to drive the motion, represented as $\tau_t\in \mathcal{R}^{N_t}$. 
$N_t$ is coupled with the number of joints $N_j$ as $N_t=N_j - 6$, with the 6-DoF free root joint unactuated.
For clarity, we denote the joint torques corresponding to Rajagopal's model as $\tau_{tr}$ and those corresponding to SMPL as $\tau_{ts}$.

\noindent{\bf Muscle Actions} represent the activation level of each muscle, resulting in the muscle forces that produce joint torques.
We denote muscle actions as $\tau_m \in \mathcal{R}^{N_m}$ with $N_m$ muscles.

\noindent{\bf sEMG} detects the electric potential generated on the body surface by activated muscle cells, which is closely related to muscle actions.
We represent sEMGs at timestamp $t$ as $\tau_e^t \in \mathcal{R}^{N_e}$ with $N_e$ sEMG channels.

\subsection{Homogeneous Dynamics Space}
\label{sec:hdys}

With the heterogeneous representations, we construct our homogeneous dynamics space (HDyS) as an aggregation of multiple auto-encoders corresponding to the inverse-forward dynamics procedure.
Then, \textit{reconstruction} and \textit{alignment} losses are adopted for supervision. 
For simplicity, all superscripts $t$ are omitted.

\noindent{\bf Inverse-Dynamics Auto-Encoder (IDAE)} encodes the kinematics into the latent space and decodes the dynamics from the latent.
For markers $x_m$ and skeletal key points $x_k$, we adopt three-layer transformer encoders with no positional embedding to encode them into latent $z_m,z_k \in \mathcal{R}^d$ of dimension $d$.
This enables the encoding of an arbitrary number of markers/points.
For joint angles $x_a$ and SMPL parameters $x_s$, we adopt simple three-layer MLPs to encode them into latent $z_a,z_s \in \mathcal{R}^d$.
The decoder is designed as a shared transformer followed by separate MLP regression heads.
The transformer takes latent vectors from consecutive frames as input, outputting \textit{per-frame} latent vectors refined with temporal contexts.
Then, separate MLP decoders decode the per-frame dynamics as $\hat{\tau}_{tr}, \hat{\tau}_{ts}, \hat{\tau}_m, \hat{\tau}_e$.

\noindent{\bf Forward-Dynamics Auto-Encoder (FDAE)} encodes dynamics and the kinematics except for accelerations (denoted as $\Tilde{x}$) into the latent space and decodes the accelerations.
Similar to IDAE, markers $\Tilde{x}_m$ and skeletal key-points $\Tilde{x}_k$ are encoded by transformers as $\Tilde{z}_m, \Tilde{z}_k$, while joint angles $\Tilde{x}_a$ and SMPL parameters $\Tilde{x}_s$ are encoded by MLPs as $\Tilde{z}_a, \Tilde{z}_p$.
Then, joint torques $\tau_{tr}, \tau_{ts}$, muscle activations $\tau_m$, and sEMGs $\tau_e$ are also encoded by separate MLP encoders as dynamics latent vectors $\Tilde{z}^{tr}, \Tilde{z}^{ts}, \Tilde{z}^m, \Tilde{z}^e$.
$\Tilde{z}_m, \Tilde{z}_k, \Tilde{z}_a, \Tilde{z}_p$ is then \textit{concatenated} with available dynamics latent vectors, fed into a shared MLP composer, resulted in latent vectors $z_{\cdot}^{\cdot}$, with subscripts representing the kinematics and superscripts representing the dynamics.
Although arbitrary combinations are feasible, in practice, only one dynamic latent vector typically exists due to the data limitations.
Finally, separate MLP decoders decode the latent vectors $z_{\{kin\}}^{\{dyn\}}$ into skeletal key point accelerations $\ddot{k}$, SMPL accelerations $\ddot{s}$, and joint angle accelerations $\ddot{a}$, where $kin \in \{m,k,s,a\}, dyn\in \{m,e,ts,tr\}$.
The marker accelerations are not predicted since their variable sizes could bring unnecessary complexity.

\noindent{\bf Loss Terms.} 
To train HDyS, we adopt reconstruction losses and alignment losses. 
For reconstruction losses, we simply calculate the L1 loss for $\tau$ and $\ddot{a}$ as 
\begin{equation}
    L_{recon} = \|\tau - \hat{\tau}\|_1 + \|\ddot{a} - \hat{\ddot{a}}\|.
\end{equation}
For alignment losses, we align the latent vectors $z$ of the same frame and separate $z$s of different frames using InfoNCE~\cite{oord2018representation}.
Given a batch of $B$ frames, the obtained latent vectors is denoted as $Z=\{z_{\{kin\}}, z_{\{kin\}}^{\{dyn\}}\}$, with $kin \in \{m,k,s,a\}, dyn\in \{m,e,ts,tr\}$.
The loss is
\begin{equation}
    L_{align} = \sum_{z_2,z_2\in Z}\sum_{i=1}^B-\log(\frac{\exp(\langle z_1^{(i)},z_2^{(i)} \rangle)}{\sum_{j=1}^B \exp(\langle z_1^{(i)},z_2^{(j)} \rangle)}),
\end{equation}
where $(i),(j)$ indicate the batch index.
The overall loss is as $\mathcal{L}=\alpha_1L_{recon}+\alpha_2L_{align}$ with coefficients $\alpha_1,\alpha_2$.
\section{Experiments}
\begin{table*}[!t]
    \centering
    \caption{Quantitative results of HDyS with ablation studies. For HDyS, the results of \textit{averaged} predictions and the \textit{best} prediction among all representations are reported.}
    \vspace{-10px}
    \resizebox{.9\linewidth}{!}{
    \begin{tabular}{lccccccc}
        \toprule
        Methods                       & ImDy                               & AddBiomechanics                 & \multicolumn{2}{c}{MinT}                              & \multicolumn{2}{c}{MiA} \\
                                      & mPJE$\downarrow$                   & mPJE$\downarrow$                & RMSE$\downarrow$          & PCC$\uparrow$                         & RMSE$\downarrow$ & PCC$\uparrow$ \\
                                      & avg/bst                            & avg/bst                         & avg/bst                   & avg/bst                               & avg/bst               & avg/bst \\
        \hline                                                                                                                        
        ImDyS~\cite{liu2024imdy}      & 0.6300                             & 0.1626                          &  -                        &  -                                    &   -                   & - \\
        MiA\cite{chiquier2023muscles} &   -                                & -                               &  -                        &  -                                    & \underline{13.3}      & - \\
        HDyS                          & \underline{0.5765}/\textbf{0.4674} & \textbf{0.1189}/\textbf{0.1243} & 0.0614/\underline{0.0615} & \underline{0.7420}/\underline{0.7402} & \textbf{11.8}/\textbf{11.6} & \textbf{0.7232}/\textbf{0.7261} \\
        \hline                                                                               
        ImDy-only HDyS                & 0.6854/0.5403                      & -                               &   -                       &   -                                   &   -                   & - \\
        AddBiomechanics-only HDyS     &  -                                 & 0.1695/0.1691                   &   -                       &   -                                   &   -                   & - \\
        MinT-only HDyS                &  -                                 & -                               & 0.0637/0.0640             & 0.7179/0.7127                         &   -                   & - \\   
        MiA-only HDyS                 &  -                                 & -                               & -                         & -                                     & 13.6/\underline{13.5} & \underline{0.6557}/0.6421 \\      
        \hline                                                                                                                        
        HDyS w/o ImDy                 &  -                                 & 0.1214/0.1386                   & \textbf{0.0608}/0.0617    & \textbf{0.7470}/\textbf{0.7417}       & 15.8/15.6             & 0.6523/0.6497 \\
        HDyS w/o AddBiomechanics      & 0.5787/0.4742                      & -                               & 0.0616/\textbf{0.0614}    & 0.7408/0.7386                         & 17.1/17.0             & 0.5769/0.5638 \\   
        HDyS w/o MinT                 & \textbf{0.5730}/\underline{0.4681} & \underline{0.1197}/0.1296       & -                         &  -                                    & 16.9/16.8             & 0.5213/0.5313 \\      
        HDyS w/o MiA                  & 0.5890/0.4788                      & 0.1200/0.1284                   & 0.0616/0.0618             & 0.7395/0.7375                         & -                     & -      \\      
        HDyS w/o AMASS                & 0.5797/0.4786                      & 0.1217/0.1319                   & \underline{0.0613}/0.0617 & 0.7419/0.7380                         & 14.7/14.9             & 0.5704/0.5632 \\      
        \hline                                
        HDyS w/o $L_{align}$          & 0.6575/0.5019                      & 0.1270/0.1329                   & 0.0626/0.0630             & 0.7318/0.7238                         & 13.7/13.4             & 0.6464/0.6402 \\       
        HDyS w/o FDAE                 & 0.5776/0.4849                      & 0.1198/\underline{0.1261}       & 0.0617/0.0617             & 0.7388/0.7375                         & 13.6/\underline{13.3}         & 0.6517/\underline{0.6699}  \\
        \hline
        HDyS-32D                      & 0.7390/0.6450                      & 0.1401/0.1420                   & 0.0650/0.0648             & 0.6980/0.7010                         & 16.7/16.3             & 0.5524/0.5614 \\
        HDyS-64D                      & 0.6505/0.5410                      & 0.1295/0.1354                   & 0.0627/0.0629             & 0.7272/0.7254                         & 15.1/14.5             & 0.6034/0.6187  \\
        \bottomrule
    \end{tabular}}
    \label{tab:res-main}
    \vspace{-15px}
\end{table*}
\subsection{Datasets}

\noindent{\bf AddBiomechanics}~\cite{werling2023addbiomechanics} contains over 50 hours of human motion data with joint torques from Nimble~\cite{werling2021fast} simulation. 
We follow the setting in \cite{werling2023addbiomechanics} with the armless part of AddBiomechanics. 
Markers, key points, and joint angles are used as kinematics representations, and joint torques corresponding to \cite{rajagopal2016full} represent the dynamics.

\noindent{\bf Muscles in Times (MinT)}~\cite{schneider2024mint} simulates part of AMASS in OpenSim to obtain the actions of 402 muscles. We randomly split them into 906 sequences for training and 227 sequences for testing. 
Markers, key points, and SMPL parameters are used as kinematics representations, and muscle actions are the dynamics representations.

\noindent{\bf Muscles in Act (MiA)}~\cite{chiquier2023muscles} consists of 12.8 hours of human exercise motion reconstructed from videos with VIBE~\cite{kocabas2020vibe} and sEMG data of 8 muscles.
Following \cite{chiquier2023muscles}, we split them into 19,563 training sequences and 3,053 testing sequences.
Markers and key points are kinematics representations, and sEMGs are dynamics representations.

\noindent{\bf ImDy}~\cite{liu2024imdy} adopted PHC~\cite{luo2023perpetual} to imitate motion sequences from AMASS~\citep{mahmood2019amass} and KIT~\citep{Mandery2016b}, resulting in over 150 hours of human dynamics data, with 27,501 sequences for training and 3,055 sequences for evaluation.
Markers, key points, and SMPL are kinematics representations, and SMPL torques are dynamics representations.

\noindent{\bf AMASS}~\cite{mahmood2019amass} contains over 11,000 motion sequences represented in SMPL~\cite{loper2023smpl}. 
Though not paired with human dynamics, we included it as an extra knowledge base for training. 
Markers, key points, and SMPL parameters are used as kinematics representations.

\subsection{Implementation Details}

We adopt a compact design for HDyS with a latent dimension of 128.
For joint angle and SMPL parameters, three-layer MLPs with hidden unit sizes of 256 and 128 are adopted as encoders, projecting the input to the 128-D latent space.
For markers and key-points, three-layer transformer encoders with 2 heads are adopted as encoders.
The transformer in the ID decoder is of 4 layers, 4 heads, and a dimension of 128.
All MLPs in decoders are two-layer, with a hidden size of 32 for Rajagopal's torque, sEMG, joint angle accelerations, and key-point accelerations, and a hidden size of 64 for SMPL torque, SMPL accelerations, and muscle actions.
For training, we adopt an AdamW optimizer with a learning rate of 1e-3 and a batch size of 9,600 frames for 1,000 epochs.
Loss weights are set as $\alpha_1=0.01, \alpha_2=0.05$.
Also, during training, a balanced sampling strategy is adopted to minimize the scale influence of different datasets.
That is, we randomly sample 3,000 sequences per dataset for each training epoch.
In this way, the comparison is fair in terms of the number of seen samples under different dataset settings.
All experiments are conducted on a single NVIDIA Titan Xp GPU.

\subsection{Results on Inverse Dynamics}
\noindent{\bf Metric. } 
We report mPJE and RMSE as 
\begin{equation}
    \label{eq:mpje}
    mPJE_{\tau_{\cdot}} = \frac{1}{J}\sum_{j=1}^{J}|\tau_{\cdot}-\hat{\tau}_{\cdot}|_2, RMSE_{\tau_{\cdot}} = |\tau_{\cdot}-\hat{\tau}_{\cdot}|_2.
\end{equation}
PCC is also reported due to its invariance against scale and offset, which helps evaluate muscle actions and sEMG prediction.
PCC is computed as 
\begin{equation}
    PCC_{\tau_{\cdot}} = \frac{cov(\tau_{\cdot}, \hat{\tau}_{\cdot})}{\sigma_{\tau_{\cdot}}\sigma_{\hat{\tau}_{\cdot}}},
\end{equation}
where $cov(\tau_{\cdot}, \hat{\tau}_{\cdot})$ is the covariance of $\tau_{\cdot}, \hat{\tau}_{\cdot}$, and $\sigma_{\cdot}$ indicates the standard deviation.
For ImDy and AddBiomechanics, we report mPJE normalized by body weight.
For MinT and MiA, RMSE and PCC are reported.

\subsubsection{Quantitative results} 
Quantitative results are shown in Tab.~\ref{tab:res-main}, where HDyS is reported with the averaged predictions from different input kinematics representations and the best predictions among all representations.
Compared to previous methods~\cite{liu2024imdy,chiquier2023muscles}, the proposed HDyS provides superior performances with substantial improvements on all datasets.
We further analyze the performance of HDyS with three questions.

\noindent{\bf How do the heterogeneous datasets contribute to HDyS?} 
We compare HDyS with its single-dataset variants and drop-one-out variants in Tab.~\ref{tab:res-main}.
HDyS outperforms the corresponding single-dataset variants on all datasets, validating the existence of homogeneous human dynamics knowledge behind these heterogeneous datasets. 
According to the drop-one-out results, it is noticeable that datasets with similar dynamics representations are more cooperative.
That is, muscle-related datasets (MiA and MinT) tend to benefit more from each other and less from torque-related datasets (AddBiomechanics and ImDy), and vice versa.
Moreover, mutually harmful effects could be observed for ImDy and MinT.
These are consistent with the gaps between these datasets: the sim2real gap from ImDy to others and the torque-to-muscle gap from AddBiomechanics and ImDy to MiA and MinT.
AMASS, though not paired with dynamics information, is shown to be beneficial for inverse dynamics with its diverse and high-quality kinematics.

The heterogeneous datasets introduce both increased data scale and heterogeneous knowledge.
Given this, we try to investigate further the source of the improvement in Tab.~\ref{tab:res-main}.
Trying to decompose the contributions of scale and heterogeneity, we compared the performance of the following three models on a single target test set in Tab.~\ref{tab:util}.
\textit{HDyS-Single-50} denotes single-dataset HDyS with 50$\%$ of the data from the target dataset.
\textit{HDyS-50/50} denotes HDyS using 50$\%$ of the data from the target dataset and 50$\%$ of the data from the other datasets, maintaining the same data scale as the target dataset. 
\textit{HDyS-Single} represents the corresponding single-dataset variants of HDyS in Tab.~\ref{tab:res-main}.
The performance of the best representation is reported on AddBiomechanics and MiA, considering their suitable scale and realistic nature.
As shown, HDyS-50/50 consistently outperforms HDyS-Single-50, verifying the existence of homogeneous knowledge in heterogeneous data.
Moreover, on AddBiomechanics, HDyS-50/50 even outperforms HDyS-Single, indicating that the diversity from heterogeneous data could sometimes benefit more than simply increasing seemingly homogeneous data.
\begin{table}[!t]
    \centering
    \caption{Results on scale-heterogeneity decomposition.}
    \vspace{-10px}
    \resizebox{\linewidth}{!}{\setlength{\tabcolsep}{3pt}
    \begin{tabular}{lccc}
        \toprule
        Dataset                          & HDyS-Single-50 & HDyS-50/50        & HDyS-Single    \\
        \hline
        AddBiomechanics $mPJE\downarrow$ & 0.1707         & \textbf{0.1284}   & 0.1695         \\
        MiA $RMSE\downarrow$             & 16.2           & 14.5              & \textbf{13.5}  \\
        \bottomrule
    \end{tabular}}
    \vspace{-10px}
    \label{tab:util}
\end{table}

\begin{table}[!t]
    \centering
    \caption{Results of different kinematic representations in HDyS.}
    \vspace{-10px}
    \resizebox{\linewidth}{!}{\setlength{\tabcolsep}{3pt}
    \begin{tabular}{lccccccc}
        \toprule
        Methods                       & ImDy               & AddBiomechanics    & \multicolumn{2}{c}{MinT}                & \multicolumn{2}{c}{MiA} \\
                                      & mPJE$\downarrow$   & mPJE$\downarrow$   & RMSE$\downarrow$   & PCC$\uparrow$      & RMSE$\downarrow$ & PCC$\uparrow$     \\
        \hline                                                
        HDyS-marker                   &  0.8163         & 0.1455          & 0.0646          & 0.6968          & 12.3          & 0.7088 \\
        HDyS-keypoint                 &  0.8084         & 0.1324          & 0.0640          & 0.7103          & \textbf{11.6} & \textbf{0.7261} \\
        HDyS-angle                    &   -             & \textbf{0.1243} &   -             &    -            &   -           & - \\   
        HDyS-SMPL                     & \textbf{0.4674} & -               & \textbf{0.0615} & \textbf{0.7402} &  -            & - \\       
        \bottomrule
    \end{tabular}}
    \label{tab:res-modal}
    \vspace{-15px}
\end{table}
\begin{figure}[!t]
    \centering
    \includegraphics[width=\linewidth]{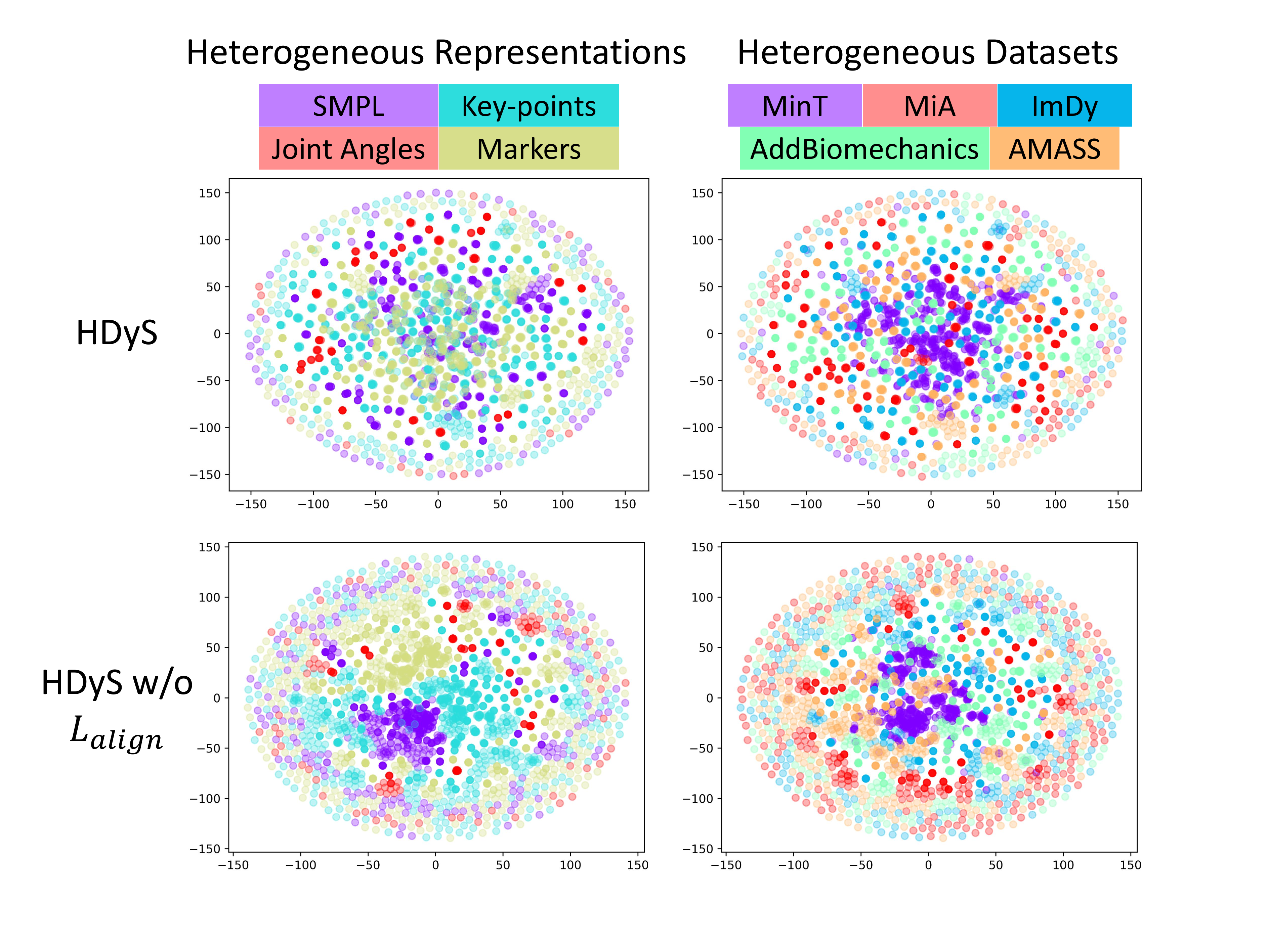}
    \vspace{-10px}
    \caption{Feature visualization for HDyS w/ and w/o $L_{align}$.}
    \vspace{-10px}
    \label{fig:feat-vis}
\end{figure}
\noindent{\bf How do the heterogeneous representations contribute to HDyS?}
We report the results from individual kinematics representations in Tab.~\ref{tab:res-modal}. 
Typically, the representations in joint space like SMPL and joint angles produce more accurate predictions compared to Cartesian representations with considerable performance gaps, which explains the performance drop when taking the average for ImDy and MiA.
The markers are the worst-performed representation of all.

In Tab.~\ref{tab:res-main}, it is noticeable that MinT and AddBiomechanics take advantage of averaging predictions from different representations. 
For AddBiomechanics, this might be because the best-performing representation is joint angles, which are not available in other datasets. 
Therefore, the averaging could effectively aggregate knowledge from other datasets with other representations.
For MinT, it is observable that different representations produce close results, preventing the ensemble from being held back by markers.

Furthermore, we examine the alignment of heterogeneous representations by removing the alignment loss $L_{align}$ in Tab.~\ref{tab:res-main}.
Considerable performance drops could be observed.
The drop is consistent for both best and average performances, indicating the cross-modality alignment manages to aggregate knowledge from heterogeneous representations.
We also visualize HDyS features w/ and w/o $L_{align}$ in Fig.~\ref{fig:feat-vis}.
As shown, $L_{align}$ prevents the HDyS feature from being dominated by the representation heterogeneity, encouraging the learning of homogeneous knowledge.
Interestingly, though not intended to do so, $L_{align}$ also helps to mitigate the gap between datasets, alleviating the domain heterogeneity.

\noindent{\bf How does the model design contribute to HDyS?}
We first evaluate the FDAE in Tab.~\ref{tab:res-main}. 
A substantial performance drop could be observed, demonstrating the benefit of informing HDyS with the forward dynamics procedure.
Moreover, we also report HDyS with different dimensions in Tab.~\ref{tab:res-main}.
Though HDyS's performance degrades with lower dimensions, it could still outperform some single-dataset variants with higher dimensions, validating the efficacy of heterogeneous datasets again.

\subsubsection{\bf Qualitative results} 
\begin{figure}[!t]
    \centering
    \includegraphics[width=\linewidth]{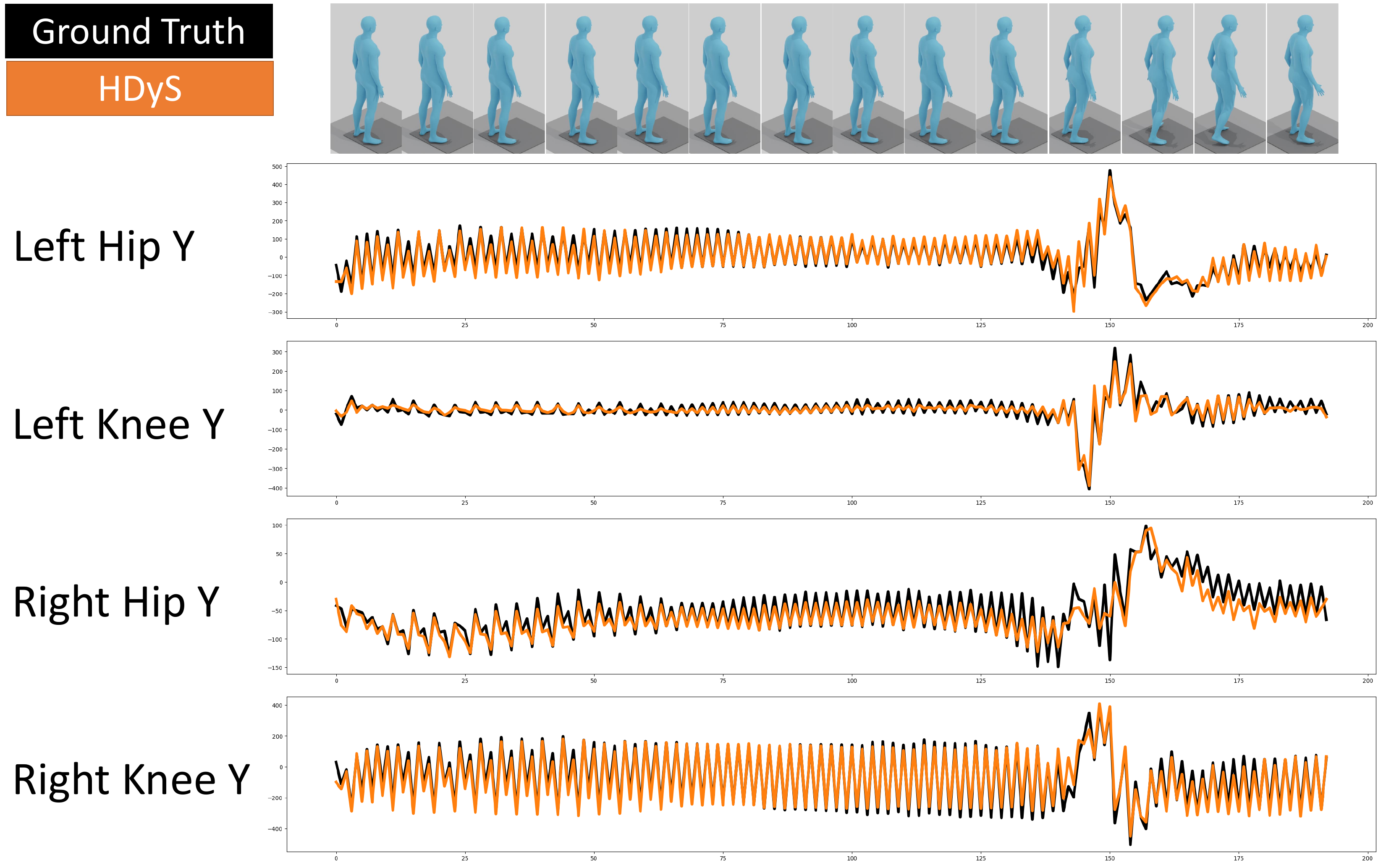}
    \vspace{-10px}
    \caption{Inverse dynamics visualization on ImDy.}
    \vspace{-15px}
    \label{fig:imdy-vis}
\end{figure}

\begin{figure}[!t]
    \centering
    \includegraphics[width=.9\linewidth]{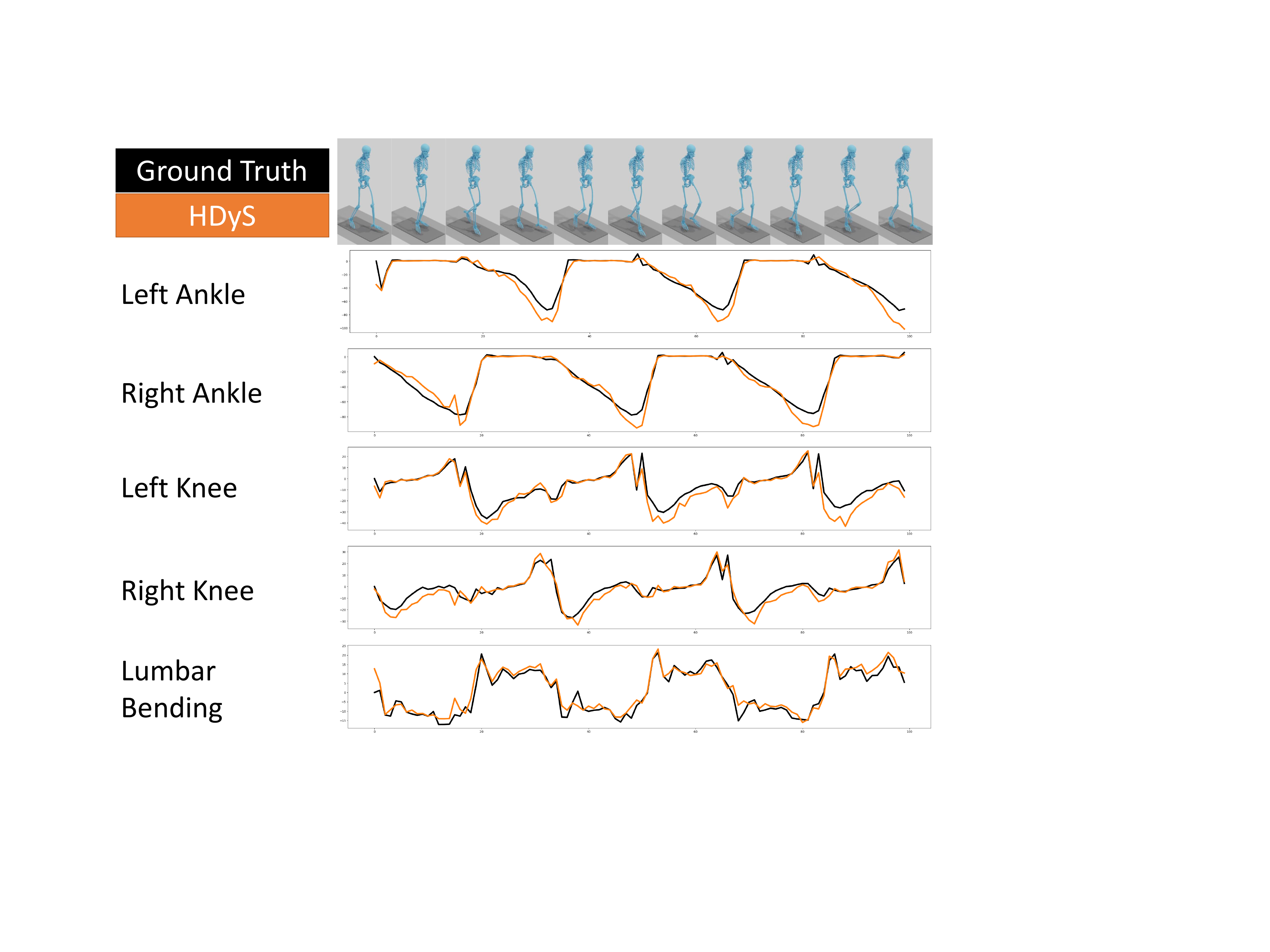}
    \vspace{-10px}
    \caption{Inverse dynamics visualization on AddBiomechanics.}
    \vspace{-15px}
    \label{fig:adb-vis}
\end{figure}

\begin{figure*}[!t]
    \centering
    \includegraphics[width=.9\linewidth]{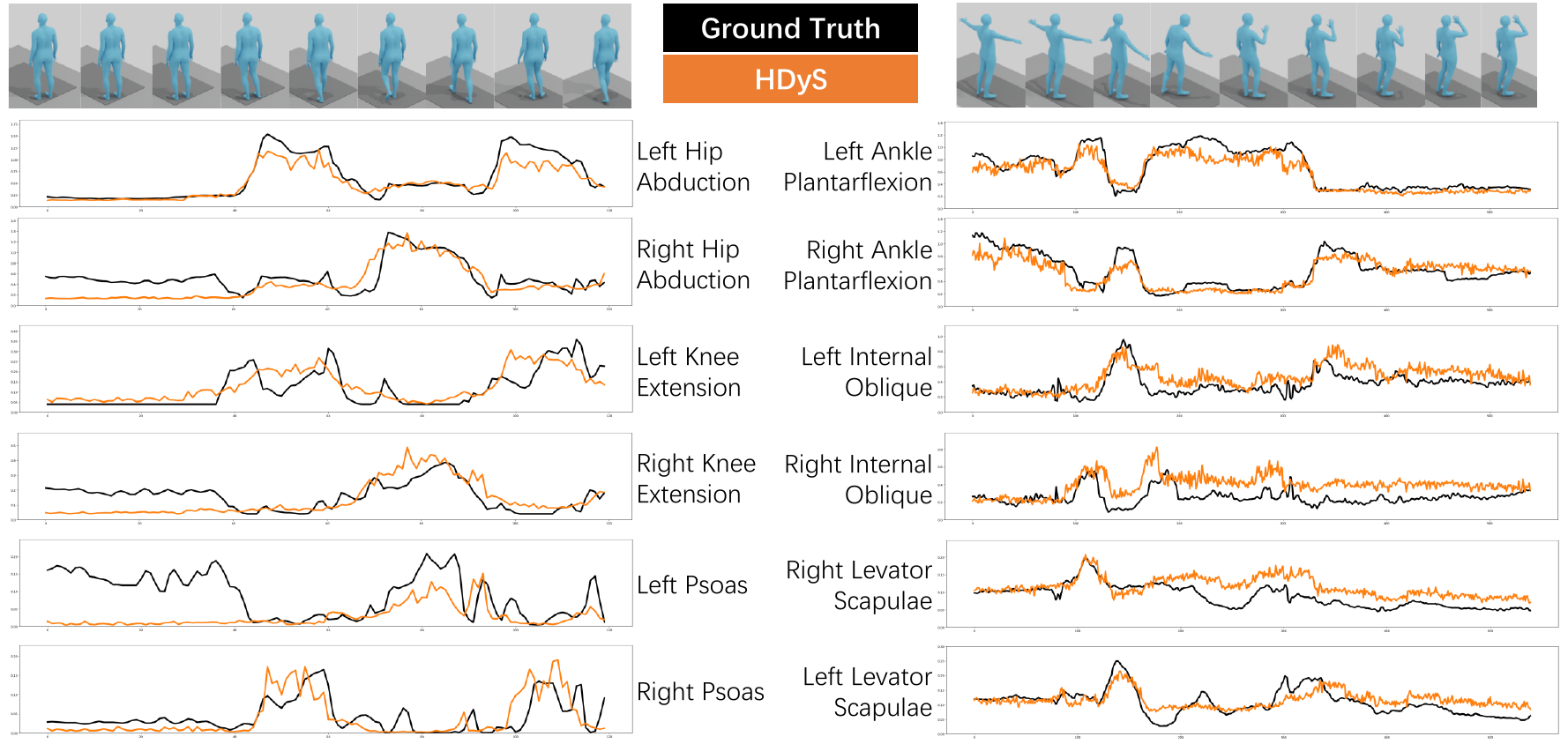}
    \vspace{-10px}
    \caption{Inverse dynamics results on MinT.}
    \label{fig:mint-vis}
    \vspace{-10px}
\end{figure*}

We visualize the inverse dynamics results of different datasets. 
As shown in Fig.~\ref{fig:imdy-vis}-\ref{fig:adb-vis}, the joint torques could be faithfully reconstructed for different actions from either synthetic and jittering ImDy or realistic AddBiomechanics with the same HDyS.
For the muscle actions shown in Fig.~\ref{fig:mint-vis}, HDyS produces reasonable estimations for lower-body muscles when walking.
Moreover, as in the right half, though the lower-body kinematics are less significant, HDyS could capture the ankle muscle actions with high fidelity. 
The predictions are also coherent with GT for upper-body muscles like the internal oblique and levator scapulae. 
More details are in the appendix.

\subsection{Results on Downstream Tasks}
\label{sec:downstream}
\subsubsection{Ground Reaction Force Estimation}
We evaluate the effectiveness of HDyS on GRF estimation with GroundLink~\cite{han2023groundlink}, which contains 1.5-hour motion with GRF recordings. We finetune HDyS with subjects 1-6 and evaluate it on subject 7. mPJE for GRF at both feet normalized by body weight is reported following Eq.~\ref{eq:mpje}.
GroundLinkNet~\cite{han2023groundlink} and HDyS trained on GroundLink from scratch are also compared.
As shown in Tab.~\ref{tab:res-glink}, the finetuned HDyS outperforms its counterparts, indicating the efficacy of the aggregated homogeneous knowledge. 
Also, the GroundLink-only HDyS outperforms GroundLinkNet, reflecting the feasibility of unifying heterogeneous representations for better dynamics knowledge.
\begin{table}[!t]
    \centering
    \caption{Results on GroundLink.}
    \vspace{-5px}
    \resizebox{\linewidth}{!}{\setlength{\tabcolsep}{3pt}
    \begin{tabular}{lccc}
        \toprule
        Methods           & GroundLinkNet & GroundLink-only HDyS & HDyS   \\
        \hline
        L-Foot $mPJE\downarrow$ & 0.0711        & 0.0584               & \textbf{0.0514} \\
        R-Foot $mPJE\downarrow$ & 0.0912        & 0.0751               & \textbf{0.0694} \\
        \bottomrule
    \end{tabular}}
    \vspace{-10px}
    \label{tab:res-glink}
\end{table}

\subsubsection{Biomechanical Human Simulation}
HDyS could also be adopted for biomechanical human simulation.
We start with the armless Rajagopal's model~\cite{rajagopal2016full} in Nimble~\cite{werling2021fast}.
Given a motion sequence, we first adopt HDyS to estimate the joint torques.
Then, we use the predicted torques to reproduce the motion in Nimble.
Starting from the current state, we feed the predicted torques for $k$ steps and compare the simulated joint angles $\hat{q}$ with the real joint angles $q$.
The simulation is performed at 90FPS.
We report the per-frame MSE of joint angles.

\begin{figure}[!t]
    \centering
    \includegraphics[width=.9\linewidth]{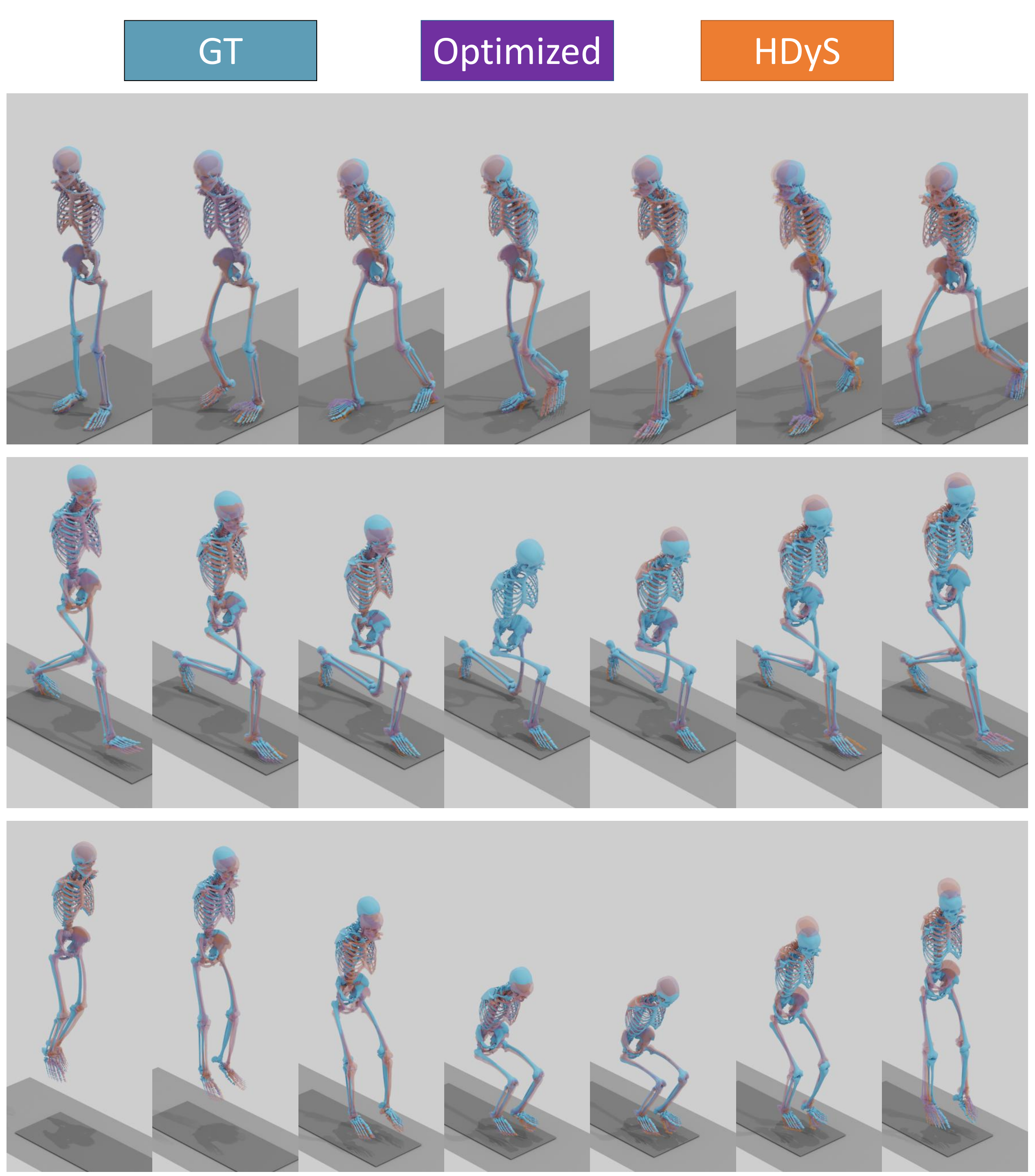}
    \vspace{-10px}
    \caption{Biomechanical human simulation visualization.}
    \vspace{-10px}
    \label{fig:fig_adb_sim}
\end{figure}

\noindent{\bf Results.} 
We demonstrated the results in Tab.~\ref{tab:nimble} and Fig.~\ref{fig:fig_adb_sim}.
The results with optimized torques are also reported as a reference.
Surprisingly, for different $k$, simulation with HDyS is superior with better stability. 
However, with $k$ increasing, MSE also increases noticeably, indicating the accumulation of drifting errors. 
Further enhancing the forward-dynamics compatibility of HDyS would be a promising goal.

\begin{table}[!t]
    \centering
    \caption{Biomechanical human simulation results reported in per-frame MSE.}
    \vspace{-5px}
    \resizebox{.9\linewidth}{!}{
    \begin{tabular}{lccccc}
        \toprule
        $k$  & 1 & 2      & 3 & 4 & 5 \\
        \hline
        Optimized & 0.00063  & 0.1909 &  1.8306 &   2.0106&  2.6027 \\
        HDyS &  \textbf{0.00061} & \textbf{0.1860} &  \textbf{1.7118} & \textbf{1.8651}  & \textbf{2.2233}  \\
        \bottomrule
    \end{tabular}}
    \vspace{-10px}
    \label{tab:nimble}
\end{table}

\subsubsection{Physical Character Control}
We adopt HDyS for physical character control following the setting of PHC~\cite{luo2023perpetual}, with 140 testing sequences and the rest for training.
We train the baseline as a primitive in PHC with a batch size of 700 for 10k steps.
Then, the HDyS latents of key points are inserted as extra observations.
We report the global and local mPJPE and the errors of velocity and acceleration in Tab.~\ref{tab:phc}.
HDyS manages to improve its performance, validating its efficacy.
\begin{table}[!t]
    \centering
    \caption{Results of HDyS for character control.}
    \vspace{-5px}
    \resizebox{.9\linewidth}{!}{
    \begin{tabular}{lcccc}
    \toprule
        Methods         & mPJPE$_g\downarrow$ & mPJPE$_l\downarrow$ & E$_{vel}\downarrow$ & E$_{acc}\downarrow$ \\ 
        \hline
        Baseline        & 74.514         & 49.029          & 11.562          & 14.243\\ 
        Baseline + HDyS &\textbf{72.692} & \textbf{47.920} & \textbf{11.305} & \textbf{13.757}\\
    \bottomrule
    \end{tabular}}
    \vspace{-10px}
    \label{tab:phc}
\end{table}

\section{Discussion}

Despite the impressive results of HDyS, it could be improved.
First, it is noticeable that HDyS performs better for the lower body than the higher body in Fig.~\ref{fig:mint-vis}.
This might be due to the imbalanced focus on lower body dynamics in data adopted by HDyS.
For all datasets, data on lower-body dynamics like gaits are \textit{dominating}. 
For AddBiomechanics, we only adopted its armless part.
Enhancing HDyS with more upper-body dynamics would be helpful.
Second, for muscle actions, HDys could omit minor changes, and the magnitudes could sometimes diverge from the real.
Mitigating it with more curated models and loss terms is desirable.
Third, as a first step toward homogeneous human dynamics learning, HDyS is primarily instantiated with five initial datasets and an intuitive model.
Scaling HDyS up with more datasets and more human priors would be a meaningful goal.
Finally, we demonstrate the potential of HDyS with some simple applications in Sec.~\ref{sec:downstream}, while more sophisticated use cases for musculoskeletal human simulation, humanoid control, and human-robot transfer learning would be promising as future works.

\section{Conclusion}
We analyzed the heterogeneity issue existing for human dynamics learning and highlighted the homogeneity beneath it.
To fully exploit the homogeneity, we proposed HDyS as a homogeneous human dynamics space.
Extensive experiments were conducted to validate the feasibility of digging homogeneity out from heterogeneity for human dynamics with detailed analyses of the contribution of heterogeneous components.
We further demonstrated the potential of HDyS for downstream applications.
We believe HDyS could shed new light on human dynamics understanding.

\section*{Acknowledgements}
This work is supported in part by the National Natural Science Foundation of China under Grant No.62306175, CCF-Tencent Rhino-Bird Open Research Fund.

{
    \small
    \bibliographystyle{ieeenat_fullname}
    \bibliography{ref}
}

\newpage
\newpage
\appendix
\section*{Appendix}
\section{Licenses}
All the data used are from the open-sourced datasets and for research purposes only. We give the links to the gathered datasets here.

\begin{itemize}
    \item AMASS: \url{https://amass.is.tue.mpg.de/license.html}
    \item Muscles in Actions: \url{https://musclesinaction.cs.columbia.edu/}
    \item AddBiomechanics: \url{https://addbiomechanics.org/download_data.html}
    \item Muscles in Time: \url{https://davidschneider.ai/mint/}
    \item ImDy: \url{https://foruck.github.io/ImDy/}
\end{itemize}

The subfigures of ``Activation Dynamics'' and ``Contraction Dynamics'' in Figure 1 are borrowed from Uchida, Thomas K., and Scott L. Delp. Biomechanics of movement: the science of sports, robotics, and rehabilitation. MIT Press, 2021. Figure 4.16 and Chapter 5.

\section{Extensive Experiments}

\subsection{Analysis on Parameters}
We compare the size of the models involved in Table 1 in Table~\ref{tab:param}. The full HDyS is comparable in \#param compared with previous efforts. In addition, it could process four heterogeneous kinematics representations and four heterogeneous dynamics representations, which could not be fulfilled with previous efforts. Moreover, even with a much smaller model scale, HDyS-32D and HDyS-64D manage to provide competitive performances, validating the efficacy of heterogeneous knowledge.

\begin{table}[!t]
    \centering
    \caption{Model size comparison.}
    \begin{tabular}{lccc}
        \toprule
        Models   &  \#params \\
        \hline   
        MiA      &  5.4M \\
        ImDyS    &  4.0M \\
        HDyS     &  3.9M \\
        HDyS-32D &  0.6M \\
        HDyS-64D &  1.4M \\
        \bottomrule
    \end{tabular}
    \label{tab:param}
\end{table}

\subsection{Extensive Results on Inverse Dynamics}
\subsubsection{Data Construction}

\begin{table*}[!t]
    \centering
    \caption{Composition of training data in Table 2.}
    \resizebox{.7\linewidth}{!}{
    \begin{tabular}{llccccc}
        \toprule
        Target dataset                  & Model         & \multicolumn{5}{c}{\#seq for training}             \\
                                        &               & AddBiomechanics & MiA     & ImDy  & MinT  & AMASS     \\ \hline
        \multirow{3}*{AddBiomechanics}  & HDyS-Single-50& 5810            & -       & -     & -     & -         \\
        ~                               & HDyS-50/50    & 5810            & 601     & 3381  & 111   & 1212      \\
        ~                               & HDyS-Single   & 11621           & -       & -     & -     & -         \\ \hline
        \multirow{3}*{MiA}              & HDyS-Single-50& -               & 2446    & -     & -     & -         \\
        ~                               & HDyS-50/50    & 526             & 2446    & 1246  & 41    & 632       \\
        ~                               & HDyS-Single   & -               & 4891    & -     & -     & -         \\
        \bottomrule
    \end{tabular}}
    \label{tab:util-appendix}
\end{table*}
To decompose the contributions of scale and heterogeneity, we construct two sets of control experiments. 
The first set of control experiments were controlled for the same data scale, and they differed only in whether the data constituted heterogeneity or not. 
The second set of control experiments varies only in the scale of the data.

Thus, we constructed HDyS-50/50 to form the first set of control experiments with the original HDyS-Single, and HDyS-Single-50 to form the second set of control experiments with HDyS-Single. In this way, HDyS-Single-50 and HDyS-50/50 formed a third control experiment with the same data from the target dataset, in which homogeneous knowledge in heterogeneous data can be observed. To construct the other datasets part of HDyS-50/50, we proportionally sampled the training data from other datasets so that the total amount of data selected was equal to 50$\%$ of the total amount of the target dataset. The details of the construction are shown in Tab.~\ref{tab:util-appendix}.

\subsubsection{More Ablation Studies}
An additional ablation study is provided to evaluate the transformer-based temporal refinement.
We remove the temporal transformer in the ID decoder and report its performance in Tab.~\ref{tab:more_ablation}.
As shown, substantial performance degradation is observed, validating the refinement of the temporal transformer.
\begin{table*}[!t]
    \centering
    \caption{Ablation study on the transformer-based temporal refinement.}
    \vspace{-5px}
    \resizebox{.9\linewidth}{!}{
    \begin{tabular}{lccccccc}
        \toprule
        Methods                       & ImDy                               & AddBiomechanics                 & \multicolumn{2}{c}{MinT}                              & \multicolumn{2}{c}{MiA} \\
                                      & mPJE$\downarrow$                   & mPJE$\downarrow$                & RMSE$\downarrow$          & PCC$\uparrow$                         & RMSE$\downarrow$ & PCC$\uparrow$ \\
                                      & avg/bst                            & avg/bst                         & avg/bst                   & avg/bst                               & avg/bst               & avg/bst \\
        \hline
        HDyS                          & \textbf{0.5765}/\textbf{0.4674} & \textbf{0.1189}/\textbf{0.1243} & \textbf{0.0614}/\textbf{0.0615} & \textbf{0.7420}/\textbf{0.7402} & \textbf{11.8}/\textbf{11.6} & \textbf{0.7232}/\textbf{0.7261} \\
        \hline        
        HDyS w/o Temporal Refinement  & 0.7002/0.5334 & 0.1393/0.1489 & 0.0666/0.0670 & 0.7372/0.7325 & 15.4/15.1 & 0.5748/0.5788 \\
        \bottomrule
    \end{tabular}}
    \label{tab:more_ablation}
\end{table*}
\subsubsection{Architectural Clarification and Justification}
Our basic idea is to use basic structures wherever possible to highlight the power of inherent homogeneity.
Therefore, we tend to use basic three-layer MLPs for single-frame fixed-size inputs (like joint angles) while maintaining non-linearity modeling ability.
Transformers are adopted when variable-size inputs (like markers and joints) or sequential inputs (in the ID decoder) are used. 
The numbers of hidden dimensions and attention heads are designed to match the dimensions of inputs/outputs.
The number of transformer layers is selected to match the number of parameters of existing baselines as listed in Appendix B.1.
While we believe HDyS could be enhanced by more sophisticated architectures like an auto-regressive operation manner, we leave this for future work.

\subsection{More Analysis on Ground Reaction Force Prediction}
In Tab.~\ref{tab:abl-glink}, we include some ablative baselines for the influence of different kinematics representations on GRF estimation, validating the mutual benefit of unifying kinematics representations again.
\begin{table}[!t]
    \centering
    \caption{More ablative baselines on GroundLink.}
    \resizebox{\linewidth}{!}{\setlength{\tabcolsep}{3pt}
    \begin{tabular}{lcccc}
        \toprule
        Methods                 & HDyS-Marker & HDyS-SMPL & HDyS-keypoint & HDyS   \\
        \hline
        L-Foot $mPJE\downarrow$ & 0.0673      & 0.0591    & 0.0584        & \textbf{0.0514} \\
        R-Foot $mPJE\downarrow$ & 0.0930      & 0.0732    & 0.1047        & \textbf{0.0694} \\
        \bottomrule
    \end{tabular}}
    \label{tab:abl-glink}
\end{table}

\subsection{More Analysis on Biomechanical Human Simulation}
Quantitative results are shown in \cref{tab:extension-sim}. 
As shown, increasing the simulation frame rate effectively reduces the simulation error. 
And HDyS consistently provides competitive performances.
However, drifting errors could still be observed. 

\begin{table}[!t]
    \centering
    \caption{Extended results reported in per-frame MSE on biomechanical human simulation.}
    \small
    \resizebox{.9\linewidth}{!}{
    \begin{tabular}{lccc}
        \toprule
        Methods           & 90FPS   & 120FPS           & 150FPS    \\
        \hline
        HDyS-2-steps      & \textbf{0.1860}  & \textbf{0.0591}           & \textbf{0.0244}         \\
        Optimized-2-steps & 0.1909  & 0.0607           & 0.0253         \\
                \hline
                
        HDyS-3-steps      & \textbf{1.7118}  & \textbf{0.5257}           & \textbf{0.2125}  \\
        Optimized-3-steps & 1.8306  & 0.5495           & 0.2223         \\
                \hline
        HDyS-4-steps      & \textbf{1.8651}  & \textbf{1.5721}           & 1.1173  \\
        Optimized-4-steps & 2.0106  & 1.7630  & \textbf{0.7081}        \\
                 \hline
        HDyS-5-steps      & \textbf{2.2233}  & \textbf{2.1384}   & \textbf{2.0482}  \\
        Optimized-5-steps & 2.6027  & 2.5147  & 2.5017         \\
        \bottomrule
    \end{tabular}}
    \label{tab:extension-sim}
\end{table}

\begin{table}[!t]
\centering
\caption{Hyperparameters for two primitives. $\sigma$: fixed variance for policy. $\gamma$:discount factor. $\epsilon$:clip range for PPO}
\label{tab:example_table}
\begin{tabular}{lccc}
\toprule
& $\sigma$ & $\gamma$ & $\epsilon$ \\ \hline
Value & 0.05 & 0.99 & 0.2 \\ 
\bottomrule
\end{tabular}
\end{table}
\subsection{Details of Physical Character Control}
We exclude all motion sequences involving sitting on chairs, walking on treadmills, leaning on tables, stepping on stairs, or floating in the air. This filtering process yields a dataset comprising 10,047 high-quality motion sequences for training and 140 sequences for testing.
Following the PHC setting, as a baseline comparison, we trained two single primitives to demonstrate that HDyS enhances physical character control performance. 
Each primitive is implemented as a six-layer MLP with units [2048, 1536, 1024, 1024, 512, 512] and employs SiLU as the activation function. 
HDyS latents corresponding to key points are incorporated as additional observations. 
The only difference between the two primitives lies in the input, one without HDyS latents denoted as \textit{Baseline}, and the other one with HDyS latents denoted as \textit{Baseline w/ HDyS}. 
For training, we employ the Adam optimizer with a learning rate of 2e-5, a batch size of 768, and train the model for 10,000 steps.
The hyperparameters used during training can be found in Table~\ref{tab:example_table}.

\end{document}